# Clustering by Constructing Hyper-Planes

**Authors:** Luhong Diao [1,2,*], Jinying Gao[1,2], Manman Deng[1,2]

**Affiliations:**

[1] Beijing Institute for Scientific and Engineering Computing, Beijing University of Technology, Beijing, China.
[2] College of Applied Sciences, Beijing University of Technology, Beijing, China
*Correspondence to: Diaoluhong@bjut.edu.cn

**Abstract:** As a kind of basic machine learning method, clustering algorithms are applied in lots of fields such as bioinformatics, marketing management, communications and pattern recognition. They group data points into different categories based on their similarity or distribution. We present a clustering algorithm by finding hyper-planes to distinguish the data points. It relies on the marginal space between the points. Then we combine these hyper-planes to determine centers and numbers of clusters. Because the algorithm is based on linear structures, it can approximate the distribution of datasets accurately and flexibly. To evaluate its performance, we compared it with some famous clustering algorithms by carrying experiments on different kinds of benchmark datasets. It outperforms other methods clearly.

Clustering algorithms reveal the intrinsic pattern of data by grouping data points into different categories. Because they always handle the datasets without pre-existing labels, only some basic information like distances between points, density of points or points' distribution can be used. Based on them, different objective functions are constructed for determining the clusters' key information like the centers of clusters (1, 2), the affinity matrix (3) or even the number of clusters (4-6).

Many algorithms require that the number of clusters is preset such as K-means (1), K-medoids (2) and algorithms based on Hierarchy (7). When the number is not given in advance, it is obviously more difficult for clustering algorithms to derive a good result. Some algorithms use the density of points to determine the number and centers of clusters, like density-based spatial clustering of applications with noise (DBSCAN) (4) and clustering by fast search and find of density peaks (CFSFDP) (5). Their problem lies in that the large difference between the densities will result in low quality performance and the results are highly sensitive to the parameters. The Affinity Propagation algorithm (6) derives the clusters' centers and number by exchanging the information between data points, which is a rather novel idea. However, it is also sensitive to the parameters while the parameter preference's value is related to the number of clusters. Some algorithms group points based on distributions (8). The performances rely on the distributions' capability to represent the data points. Therefore they cannot present different kinds of manifolds flexibly.

Here, we present a clustering method by constructing hyper-planes. It has its basis in an assumption that one group can be divided into subgroups the points of which lie in a locally linear manifold. Therefore the appropriate hyper-planes are found to distinguish points in different subgroups. By combining these hyper-planes, some connective components can be derived. Then the clusters' number and centers can be determined based on these connective components. One of its main contributions lies in that it can find the number of clusters automatically. The other is that it depends only on the marginal space between the points in different clusters. Therefore it can be approached for the data points which take the distribution on complex manifolds.

Suppose a data set contains a group of points such that $x \in S$. The $S$ can be grouped into two sets $S_1$ and $S_2$ by using K-means method. Define $Label(x) = 0$ if $x \in S_1$ and $Label(x) = 1$ if $x \in S_2$. With these labels a hyper-plane $l:(w,b)$ can be found based on Support Vector Machine (SVM) (9). Then $S$ is grouped into two categories $S'_l = \{x \mid w \bullet x + b \geq 0\}$ and $S''_l = \{x \mid w \bullet x + b < 0\}$. The points in two sets $S'_l$ and $S''_l$ are defined as affiliated to $l$. The same operation is done both on $S'_l$ and $S''_l$ recursively until data sets are smaller than a threshold $\delta$. After that, a group of hyper-planes, denoted as $L$, are derived. The map from $S$ to $L$ is defined as $f_\delta : S \to L$.

To select the proper hyper-planes automatically from $L$, $L$ is grouped into two categories based on the value of $\|w\|$ of hyper-planes by using the K-means method. The category with smaller $\|w\|$, denoted as $L'$, is the hyper-planes which prefer to distinguish the points well just as the eight lines in Fig.1A. The map from $L$ to $L'$ is defined as $\varphi : L \to L'$.

For dealing with the more complex manifolds as shown in Fig.1B and Fig.1C, $S$ is divided into some subgroups $S_{L',1}, S_{L',2}..., S_{L',m}$ such that points in $S_{L',i}, 1 \leq i \leq m$ lie on same side of each hyper-plane in $L'$. Then a new hyper-plane set $\varphi\left(\bigcup_{1 \leq i \leq m} f_\delta(S_{L',i})\right)$ can be derived. If $L'$ is not equal to $L' \cup \varphi\left(\bigcup_{1 \leq i \leq m} f_\delta(S_{L',i})\right)$, let $L' = L' \cup \varphi\left(\bigcup_{1 \leq i \leq m} f_\delta(S_{L',i})\right)$. Repeat this operation until $L'$ is unvaried. The lines in Fig. 1B and Fig. 1C are the $L'$. These three distributions in Fig.1 are the most common ones for benchmarking clustering algorithms.

Then an affinity matrix can be determined based on a hyper-plane set $H$. If

$|w \cdot x_i + b| > 1/\|w\|$ or $x$ is affiliated to $l$, $x$ is defined as isolated to $l$. If two points $x_i$ and $x_j$ lie opposite side of any hyper-plane $l$ in $H$ and one of them is isolated to $l$, $x_i$ and $x_j$ are not adjacent. Otherwise, they are adjacent. Therefore the connective components based on the affinity matrix can be derived. We define this map is $\tau : H \to \{C_1, C_2, ..., C_k\}$, where $C_i, 1 \leq i \leq k$ is a connective component.

Although sometimes the connective components are the perfect clusters as shown in Fig.1, it is not always so satisfying. So the connective components aren't viewed as clusters directly. Before finding the clusters, we firstly compute a measure of mean intra-connective-component distance and nearest-connective-component distance for each connective component as following:

The points $x$ are firstly transformed into $x^{(H)} = (d^L(x, l_1), ..., d^L(x, l_{|H|}))$ where $l_i \in H$ $(1 \leq i \leq |H|)$ and $d^L(x, l) = (wx + b)/\|w\|$. Then the mean intra-cluster distance of $C_i$ is computed as:

$$IntraDis(i) = \sum_{x_p, x_q \in C_i} \frac{2 d_g(x_p^{(H)}, x_q^{(H)})}{|C_i|(|C_i| - 1)} \quad (1)$$

where $d_g(x_i, x_j)$ is the geodesic distance between $x_i$ and $x_j$ based on the affinity matrix. The nearest-connective-component distance of $C_i$ is computed as:

$$InterDis(i) = \frac{1}{|C_i|} \sum_{x_p \in C_i} \min_{x_q \notin C_i} d(x_p^{(H)}, x_q^{(H)}) \quad (2)$$

where $d(x_i, x_j)$ is the Euclidean distance between $x_i$ and $x_j$. Based on them, the measure $M$ of $C_i$ is:

$$M_i = \begin{cases} \dfrac{InterDis(i)}{IntraDis(i)} & |C_i| > \max(1, \dfrac{\delta}{4}) \\ 0 & |C_i| \leq \max(1, \dfrac{\delta}{4}) \end{cases} \quad (3).$$

The threshold $\dfrac{\delta}{4}$ is set to filter small size connective components off. Points in connective component with the larger $M$ prefer to be in the same cluster with higher probability.

If $x_p \in C_i$, $x_q \in C_j$ and $d(x_p^{(H)}, x_q^{(H)}) = \min_{x \notin C_i} d(x_p^{(H)}, x^{(H)})$, $C_j$ is a neighbor of $C_j$. If $M_i$ is larger than its neighbors' $M$, $C_i$ is a peak connective components (PCC). If only one neighbor's $M$ is larger than $M_i$, $C_i$ is called hillside connective component (HCC). PCC prefers to be the best connective component among all its adjacent ones. So it is like a center. To make the result more accurate, we adopt the PCC whose $M$ is larger than half of the connective components as a cluster center.

Then we can derive a cluster $R_i$ whose center is $C_i$. Put $C_i$ into $R_i$. And repeat the following operation until no connective component in $R_i$ is unvisited. Pick an unvisited connective component $C_p$ from $R_i$. If $C_q$ is a neighbor of $C_p$, $M_p > M_q$ and $C_q$ is a HCC, put $C_q$ into $R_i$. After all such $C_q$ are done, mark $C_p$ is visited.

For benchmarking our algorithm, we applied our algorithm on some real-world datasets. All the data points are centralized by subtracting their mean before processing. And each hyper-plane dataset $H$ in these experiments is set to be $L' \cup \left( \bigcup_{1 \leq i \leq m} f_\delta(S_{L',i}) \right)$. We also present a simple method to determine the parameter $\delta$ automatically. Because the quantity of points put in the clusters is related to the value of $\delta$, we can draw a line chart for their relationship as shown in Fig.S. The value of X-axis and Y-axis are $\delta$ and quantity of the points put in clusters $N(\delta)$, respectively. Then we find the first peak whose $N(\delta)$ is larger than half of quantity of total points. If there is no such peak, we use the highest peak in the chart. X-axis value of the point just before this peak is set to be $\delta$ which is used for clustering. The experiments' results as shown in Fig.S indicate that this method is effective although it doesn't obtain the best results always. After $\delta$ is determined, the algorithm is repeated 5 times to derive the best result.

The first four benchmark datasets are obtained from the University of California Irvine Machine Learning Repository (10). The data in one dataset is viewed as a matrix. Row vectors represent data points. Because the different columns have different meanings, values of each column are normalized by using min-max normalization. Table.1 contains a summary of these datasets. Whereas the DBSCAN and Affinity Propagation are sensible to the parameters' values, the algorithm is also compared with the K-means (best result by repeating 20 times), Agglomative clustering algorithm and balanced iterative reducing and clustering using hierarchies

(BIRCH). These five algorithms are implemented by using the Scikit-learn library in Python and all the parameters' values are default ones.

The performances are measured by two benchmark measures Adjusted Rand Index (ARI) (11) and Normalized Mutual information (NMI) (12) as shown in Table.2 and Table.3. Because we only put part of the points into clusters, the results on these points in clusters are also presented in Table.4 and Table.5.

Rodriguez and Laio have pointed that the Olivetti Face Database (13) poses a serious challenge for algorithms to find the number of clusters automatically because the "ideal" number of clusters is comparable with the number of elements in the data set (namely of different images, 10 for each subject) (6). So we approached the algorithm on it. The results of CFSFDP and our algorithm both contained no single cluster included images of two different persons. We selected 26 clusters correctly as shown in Fig.2A while CFSFDP got 22 correct clusters (6). Because another benchmark dataset Yale Face Dataset (YFD) (14) also contains the clusters the number of which is comparable with the images' number for one person, we applied our algorithm on it as shown in Fig.2B. There are images of different persons put into one cluster. Those clusters are cluster 6 and cluster 19. The number is the order of its center as to the measure $M$. The performances on both datasets measured by ARI and NMI are also included in Table.2, Table.3, Table.4 and Table.5. Our algorithm outperforms other algorithms clearly.

To find data's intrinsic patterns, one common used idea is to approximate nonlinear models by using linear ones. Our algorithm is based on this idea. By combining the linear structures, the final results can approximate the nonlinear manifolds more flexibly and accurately.

**References and Notes:**
1. J. MacQueen, Some methods for classifification and analysis of multivariateobservations, *Proceedings of the Fifth Berkeley Symposium on Mathematical Statistics and Probability*, 281-297 (1967).
2. L. Kaufman, P. J. Rousseeuw, "Clustering by means of medoids" in Dodge Y, editor. *Statistical data analysis based on the L1 norm and related methods* (Y. Dodge, Amsterdam, ed. 1987), pp. 405-416.
3. A.Y. Ng, M.I. Jordan, Y. Weiss. On Spectral Clustering: Analysis and an Algorithm. *Adv Neural Inf Proc Syst*. MIT Press. (2001).
4. M. Ester, H.P. Kriegel, J. Sander, X. Xu, A density-based algorithm for discovering clusters in large spatial databases with noise, *Proceedings of the Second International Conference of Knowledge Discovery and Data Mining*, 323–333 (1996).
5. A. Rodriguez, A. Laio, Clustering by fast search and find of density peaks. *Science* **344**, 1492-1496 (2014).
6. B. J. Frey, D. Dueck, Clustering by Passing Messages Between Data Points. *Science*.**315**, 972-976(2007)
7. T. Zhang, R. Ramakrishnan, M. Livny, BIRCH: an efficient clustering method for very large database, *ACM SIGMOD Record. **25***, 103-114 (1996).
8. G. J. McLachlan, T. Krishnan, The EM Algorithm and Extensions (Wiley Series in

Probability and Statistics vol. 382, Wiley-Interscience, New York, 2007).
9. B.E.Boser, I.M. Guyon, V.N.Vapnik, A Training Algorithm for Optimal Margin Classifiers. *Proceedings of the 5th Annual Workshop on Computational Learning Theory*, Pittsburgh, 144-152(1992).
10. National Science Foundation, University of California Irvine Machine Learning Repository. *https://archive.ics.uci.edu/ml/index.php*, 2018-9-24/2020-4-17.
11. L. Hubert, P. Arabie, Comparing Partitions, *Journal of Classification.* **2**, 193-218 (1985).
12. M. Meila. Comparing clusteringsan information based distance. *Journal of Multivariate Analysis*, 98, 873–895 (2007).
13. F. S. Samaria, A. C. Harter, Parameterisation of a stochastic model for human face identification, *Proceedings of 1994 IEEE Workshop on Applications of Computer Vision*, pp. 138–142.

14. Yale University, Yale face database. *http://cvc.cs.yale.edu/cvc/projects/yalefaces/yalefaces.html*, 1997-9-10/2020-4-17.

**Acknowledgements:**

We acknowledge financial supports from National Natural Science Foundation of China (Grants 11772013) and general research projects of Beijing Educations Committee in China (Grants KM201910005013).

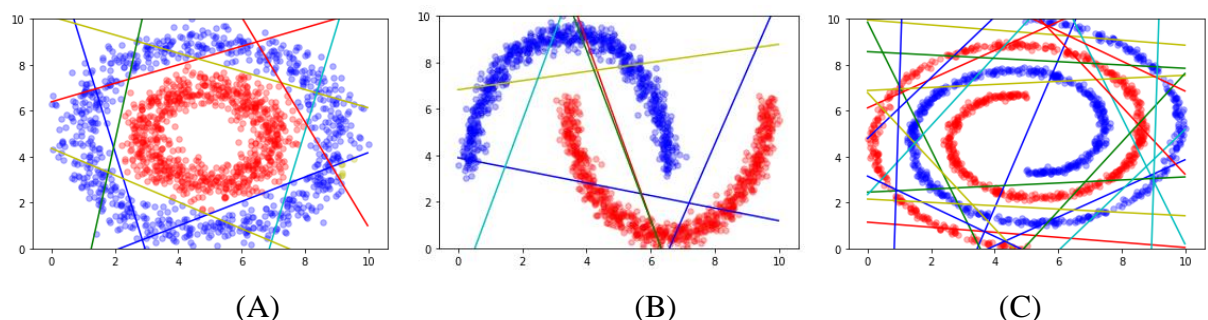

Fig.1. Results of connective components for synthetic points distribution. Different colors represent different connective components. The hyper-plane set $H = L'$. The colored lines belong to $H$. The number of points is 1600. (A) Distribution of two concentric circles. $\delta = 110$. (B) Distribution of two interleaving half circles. $\delta = 110$. (C) Distribution of swiss-roll. $\delta = 45$.

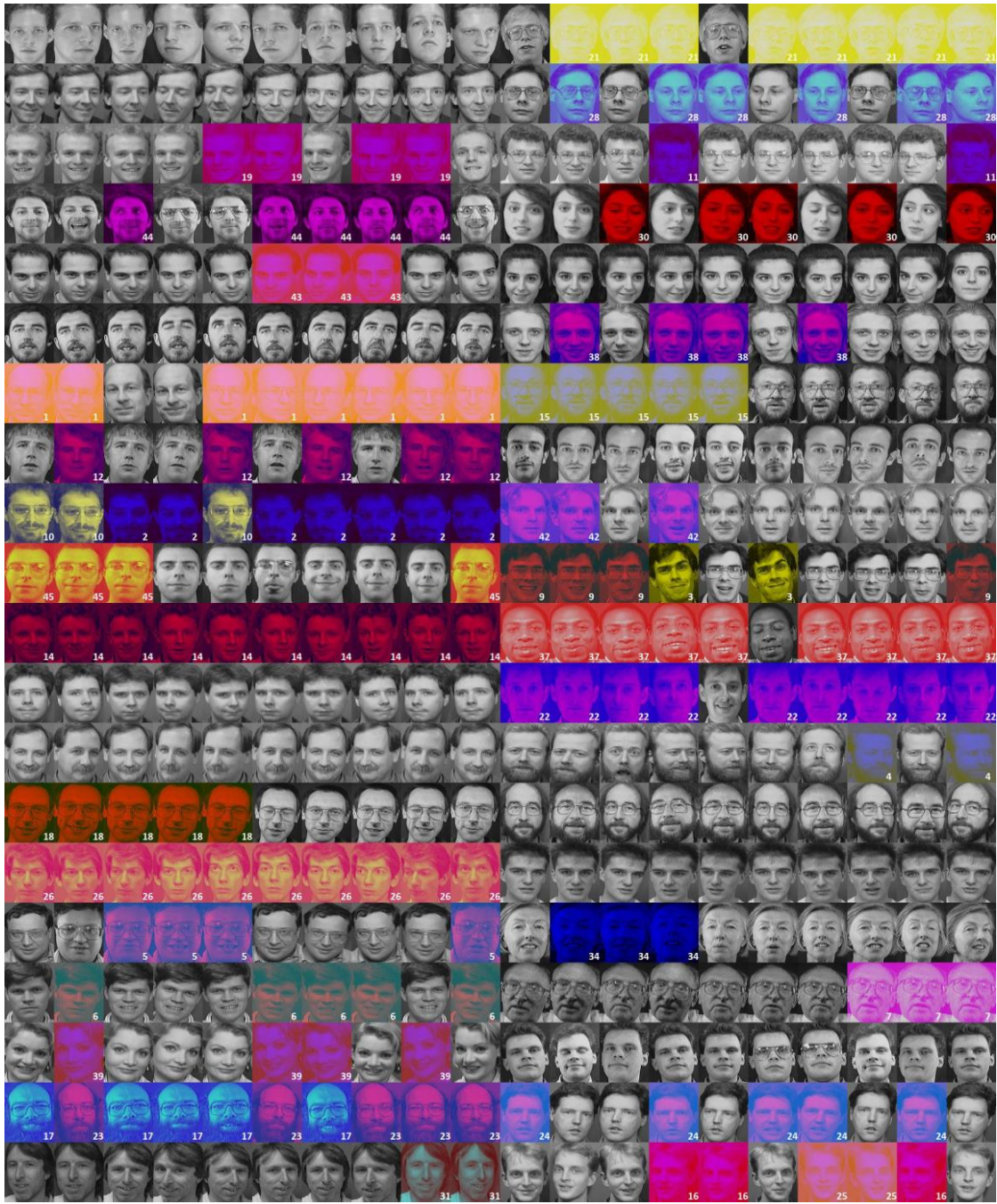

(A)

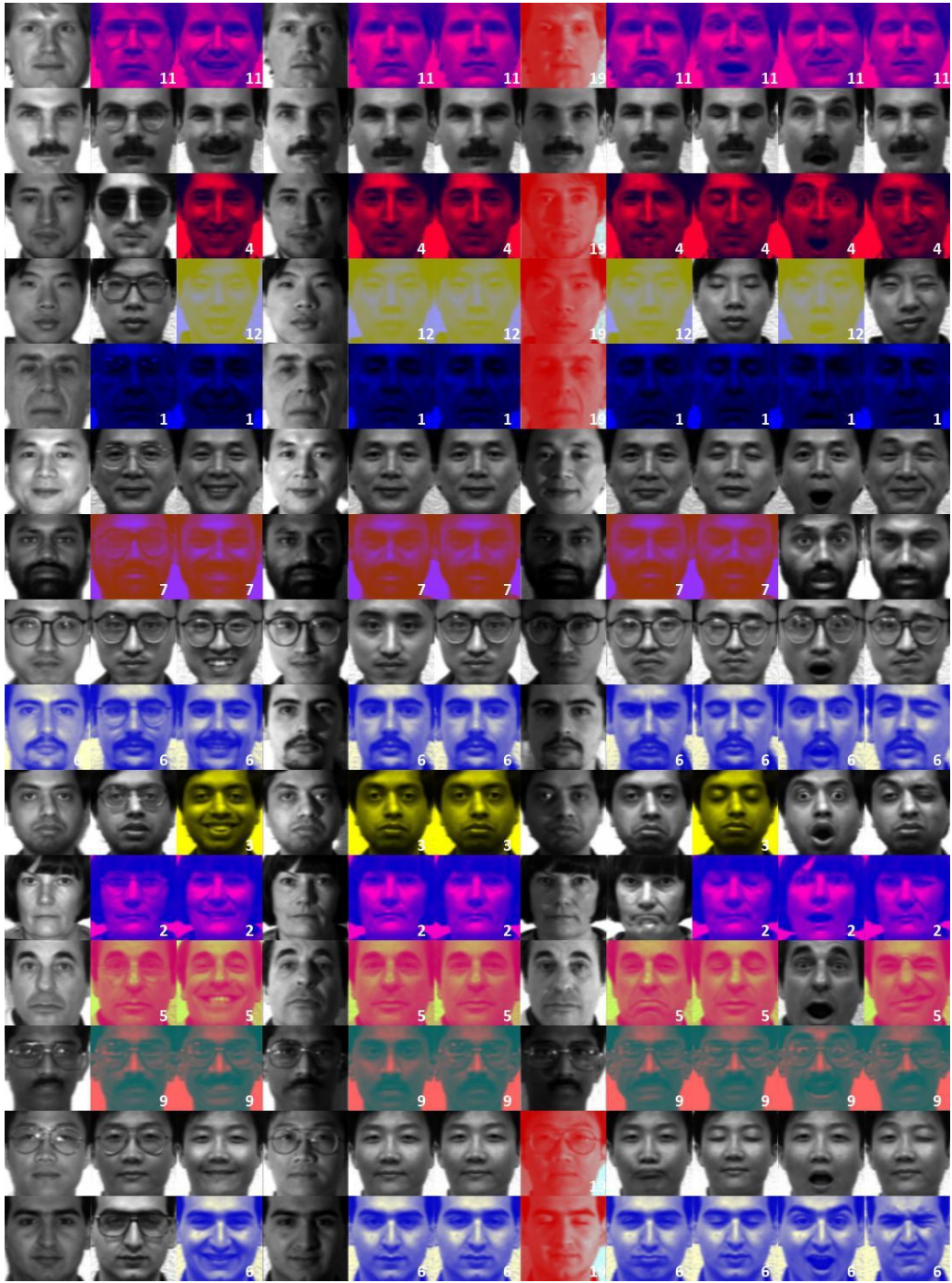

(B)

Fig.2. Pictorial cluster results on Olivetti Face Dataset (A) and Yale Face Dataset (B). Faces with same color and number in bottom-right corner belong to the same cluster while the gray images are not in any cluster. Numbers in bottom-right corner are the orders of cluster centers sorted as to their $M$.

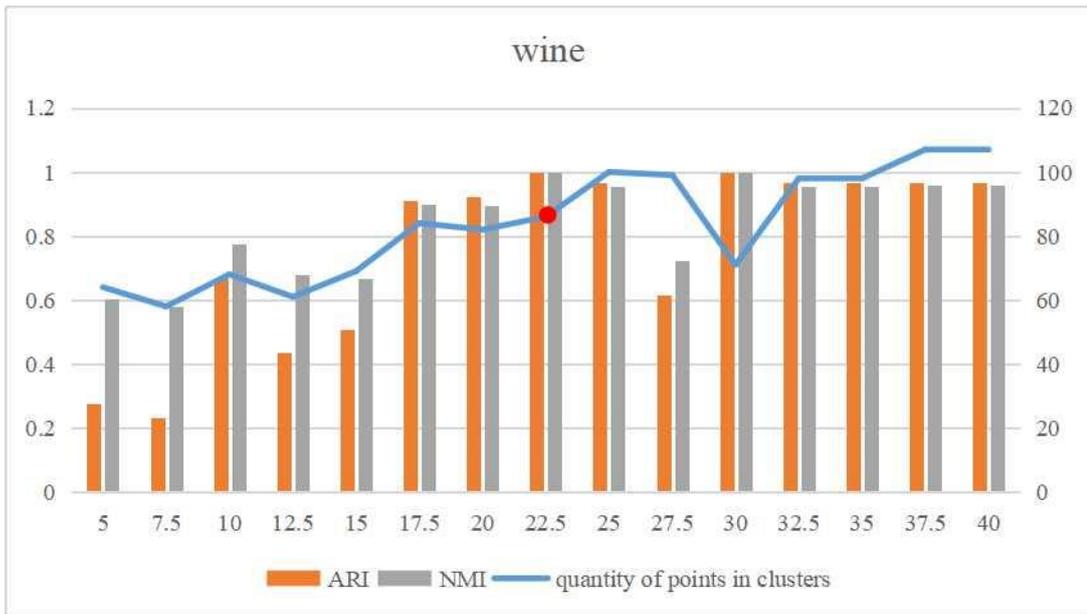

(A)

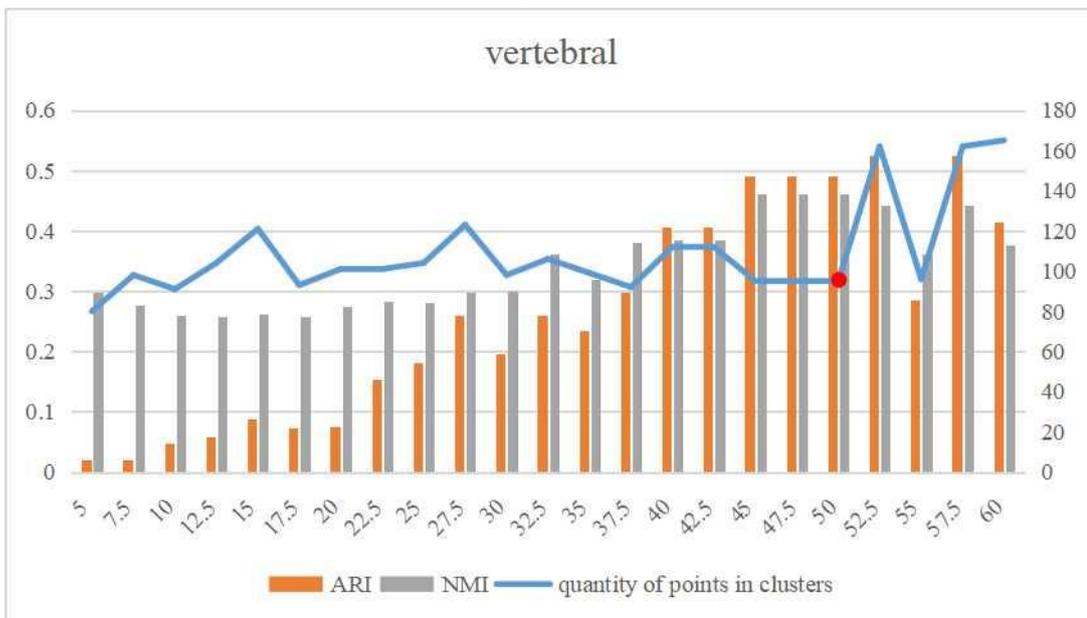

(B)

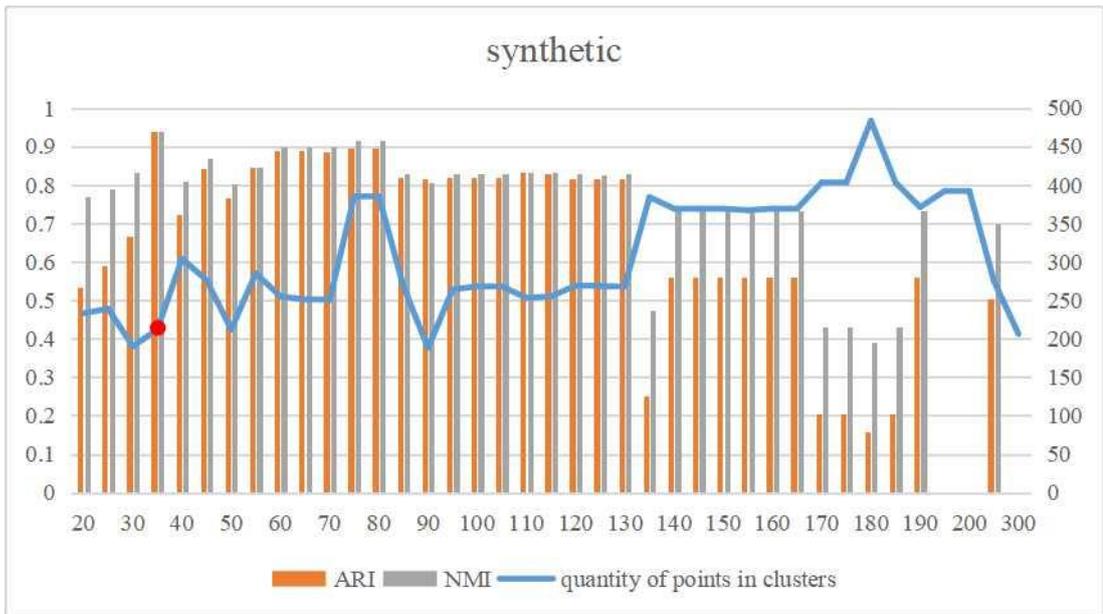

(C)

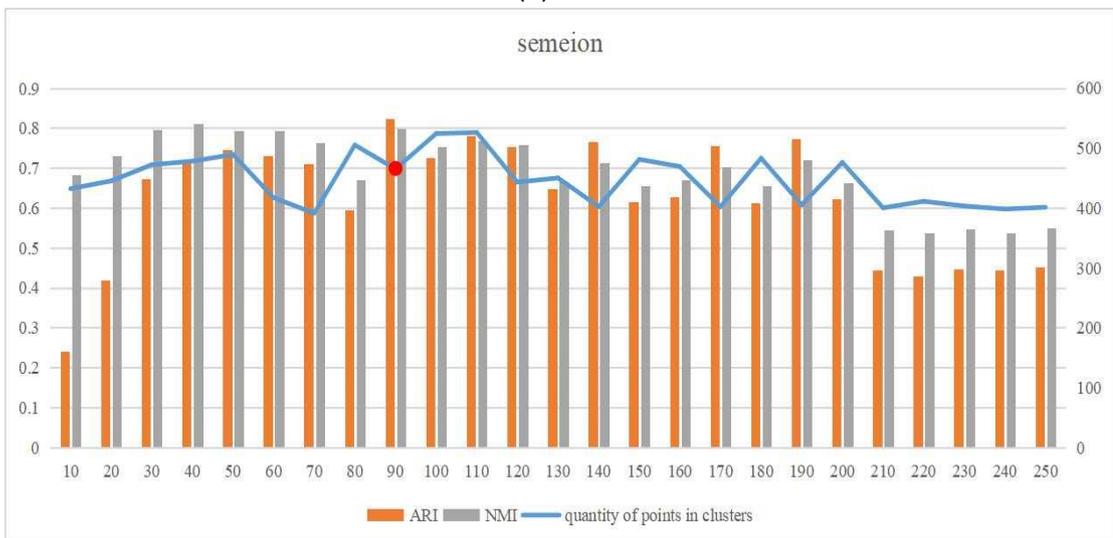

(D)

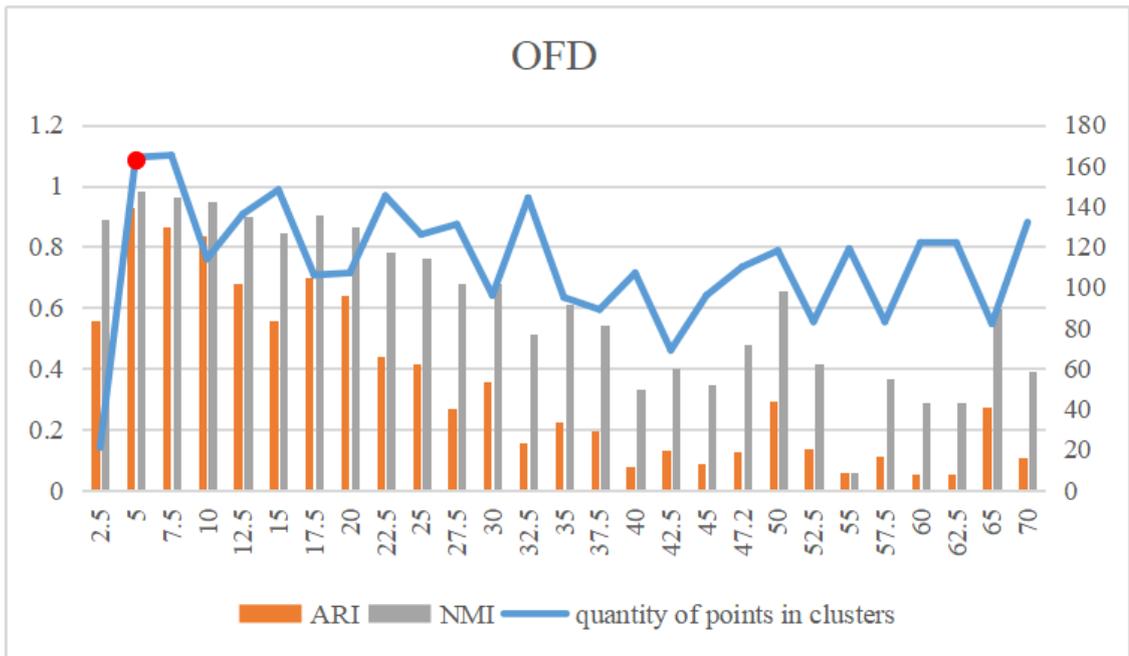

(E)

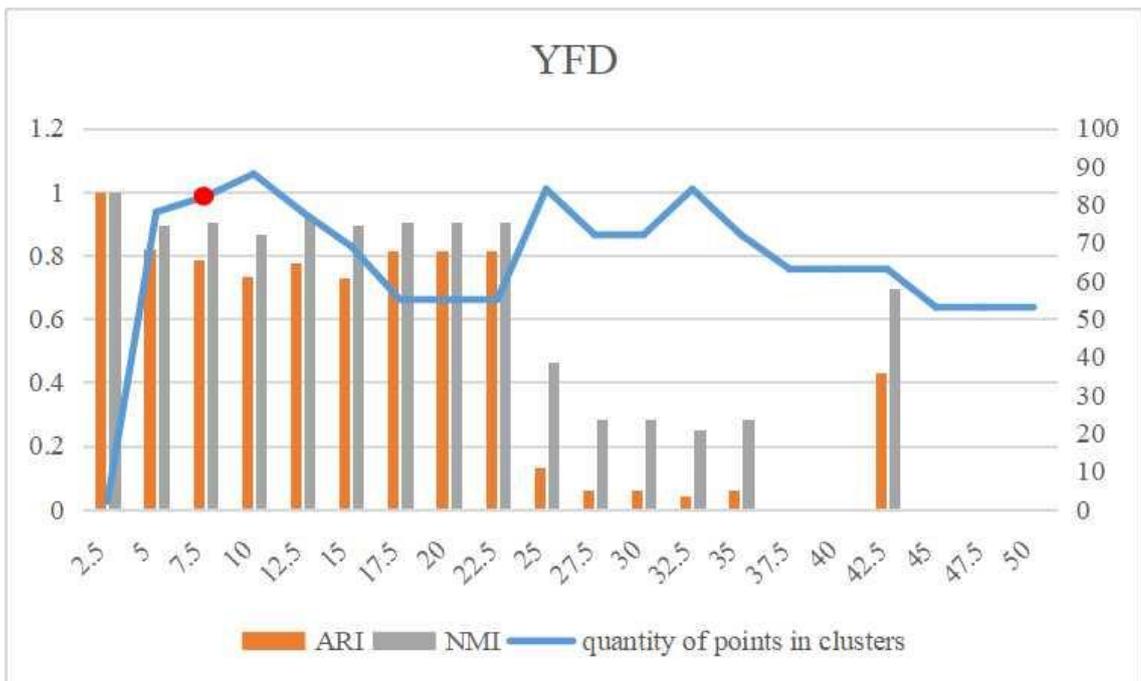

(F)

**Fig. S.**

The Performance of the algorithm on the benchmark datasets as a function of $\delta$: ARI (orange bar), NMI (gray bar) and quantity of points in clusters (blue line). The red spot is the value of $\delta$ which is selected for clustering algorithm. Numbers of cluster centers from (A) to (F) are 3, 6, 4, 5, 34 and 11, respectively.

Table 1 Descriptions of 6 Benchmark Datasets.

| Datasets | Num of Instances | Dimensions | Classes |
|---|---|---|---|
| Wine | 178 | 13 | 3 |
| Vertebral | 130 | 6 | 3 |
| Synthetic | 600 | 60 | 6 |
| Semeion | 1593 | 256 | 10 |
| OFD | 400 | 112×92 | 40 |
| YFD | 165 | 100×100 | 40 |

Table 2. Clustering ARI on the Real-world Datasets

| Datasets | K-Means | Agglomerative | Affinity Propagation | DBSCAN | BIRCH | Our algorithm |
|---|---|---|---|---|---|---|
| Wine | 0.87 | 0.93 | 0.27 | 0 | 0.93 | **1** |
| Vertebral | 0.21 | 0.24 | 0.06 | 0 | 0.10 | **0.49** |
| Synthetic | 0.63 | 0.62 | 0.38 | 0 | 0.57 | **0.94** |
| Semeion | 0.44 | 0.42 | 0.10 | 0 | 0.20 | **0.71** |
| OFD | 0.65 | 0.69 | 0.65 | 0 | 0.07 | **0.93** |
| YFD | 0.47 | 0.44 | 0.57 | 0 | 0.02 | **0.79** |

Table 3. Clustering NMI on the Real-world Datasets

| Datasets | K-Means | Agglomerative | Affinity Propagation | DBSCAN | BIRCH | Our algorithm |
|---|---|---|---|---|---|---|
| Wine | 0.85 | 0.91 | 0.53 | 0 | 0.91 | **1** |
| Vertebral | 0.27 | 0.27 | 0.25 | 0 | 0.23 | **0.46** |
| Synthetic | 0.78 | 0.81 | 0.66 | 0 | 0.76 | **0.94** |
| Semeion | 0.57 | 0.59 | 0.54 | 0 | 0.40 | **0.76** |
| OFD | 0.88 | 0.90 | 0.88 | 0 | 0.41 | **0.98** |
| YFD | 0.69 | 0.67 | 0.75 | 0 | 0.14 | **0.91** |

Table 4. Clustering ARI on Points in Clusters of Real-world Datasets

| Datasets | K-Means | Agglomerative | Affinity Propagation | DBSCAN | BIRCH | Our algorithm |
|---|---|---|---|---|---|---|
| Wine | 1 | 1 | 0.57 | 0 | 1 | 1 |
| Vertebral | **0.52** | 0.51 | 0.22 | 0 | 0.23 | 0.49 |
| Synthetic | 0.77 | 0.77 | 0.39 | 0 | 0.83 | **0.94** |
| Semeion | 0.54 | 0.47 | 0.18 | 0 | 0.60 | **0.71** |
| OFD | 0.80 | 0.72 | 0.72 | 0 | 0.13 | **0.93** |
| YFD | **0.84** | 0.80 | **0.84** | 0 | 0.04 | 0.79 |

Table 5. Clustering NMI on Points in Clusters of Real-world Datasets

| Datasets | K-Means | Agglomerative | Affinity Propagation | DBSCAN | BIRCH | Our algorithm |
|---|---|---|---|---|---|---|
| Wine | **1** | **1** | 0.76 | 0 | **1** | **1** |
| Vertebral | 0.48 | **0.49** | 0.36 | 0 | 0.46 | 0.46 |
| Synthetic | 0.83 | 0.84 | 0.69 | 0 | 0.85 | **0.94** |
| Semeion | 0.69 | 0.69 | 0.58 | 0 | 0.67 | **0.76** |
| OFD | 0.95 | 0.94 | 0.91 | 0 | 0.50 | **0.98** |
| YFD | **0.91** | 0.89 | **0.91** | 0 | 0.25 | **0.91** |